\pdfoutput=1
\documentclass[twoside,11pt]{article}

\usepackage{jmlr2e}
\usepackage{natbib}
\usepackage{algorithm}
\usepackage{algorithmic}
\usepackage{amsmath}
\usepackage{graphicx}
\usepackage{booktabs}       
\usepackage{tablefootnote}
\usepackage{footmisc}

\jmlrheading{1}{2000}{1-48}{4/00}{10/00}{Itay Hubara\u{a}}

\ShortHeadings{Quantized Neural Networks}{Hubara, Courbariaux, Soudry, El-Yaniv and Bengio} 
\firstpageno{1}

\begin{document}

\title{Quantized  Neural Networks: Training Neural Networks with Low Precision  Weights and Activations}

\author{\name Itay Hubara* \email itayh@campuse.technion.ac.il \\
       \addr Department of Electrical Engineering\\
       Technion - Israel Institute of Technology \\
       Haifa, Israel
       \AND
       \name Matthieu Courbariaux* \email matthieu.courbariaux@gmail.com \\
       \addr Department of Computer Science and Department of Statistics\\
       Universit\'{e} de Montr\'{e}al\\
       Montr\'{e}al, Canada
       \AND
       \name Daniel Soudry \email daniel.soudry@gmail.com \\
       \addr Department of Statistics\\
       Columbia University\\
       New York, USA
       \AND
       \name Ran El-Yaniv \email rani@cs.technion.ac.il \\
       \addr Department of Computer Science \\
       Technion - Israel Institute of Technology \\
       Haifa, Israel
       \AND
       \name Yoshua Bengio \email yoshua.umontreal@gmail.com \\
       \addr Department of Computer Science and Department of Statistics\\
       Universit\'{e} de Montr\'{e}al\\
       Montr\'{e}al, Canada\\\\
       *Indicates first authors.
       }

\editor{}

\maketitle

\begin{abstract}
We introduce a method to train Quantized Neural Networks (QNNs) --- neural networks with extremely low precision (e.g., 1-bit) weights and activations, at run-time. At train-time the quantized weights and activations are used for computing the parameter gradients. During the forward pass, QNNs drastically reduce memory size and accesses, and replace most arithmetic operations with bit-wise operations. As a result, power consumption is expected to be drastically reduced. We trained QNNs over the MNIST, CIFAR-10, SVHN and ImageNet datasets. The resulting QNNs achieve prediction accuracy comparable to their 32-bit counterparts. For example, our quantized version of AlexNet with 1-bit weights and 2-bit activations achieves $51\%$ top-1 accuracy. Moreover, we quantize the parameter gradients to 6-bits as well which enables gradients computation using only bit-wise operation. Quantized recurrent neural networks were tested over the Penn Treebank dataset, and achieved comparable accuracy as their 32-bit counterparts using only 4-bits. Last but not least, we programmed a binary matrix multiplication GPU kernel with which it is possible to run our MNIST QNN 7 times faster than with an unoptimized GPU kernel, without suffering any loss in classification accuracy. The QNN code is available online.
\end{abstract}


\section{Introduction}
Deep Neural Networks (DNNs) have substantially pushed Artificial Intelligence (AI) limits in a wide range of tasks, including but not limited to object recognition from images~\citep{Krizhevsky-2012-small,Szegedy-et-al-arxiv2014},
speech recognition~\citep{Hinton-et-al-2012,Sainath-et-al-ICASSP2013},
statistical machine translation~\citep{Devlin-et-al-ACL2014,Sutskever-et-al-NIPS2014,Bahdanau-et-al-ICLR2015-small},
Atari and Go games \citep{Mnih-et-al-2015,Silver-et-al-2016},
and even computer generation of abstract art \citep{Mordvintsev-et-al-2015}.

Training or even just using neural network (NN) algorithms on conventional
general-purpose digital hardware (Von Neumann architecture)
has been found highly inefficient due to the massive amount of multiply-accumulate operations (MACs) required to compute the weighted sums of the neurons' inputs. Today, DNNs are almost exclusively trained on one or many very fast and power-hungry Graphic Processing Units (GPUs) \citep{Coates-et-al-2013}. As a result, it is often a challenge to run DNNs on target low-power devices, and substantial research efforts are invested in speeding up DNNs at run-time on both general-purpose \citep{Vanhoucke-et-al-2011,Gong-et-al-2014,Romero-et-al-2014,Han-et-al-2015} and specialized computer hardware \citep{Farabet-et-al-2011-a, Farabet-et-al-2011-b, Pham-et-al-2012, Chen-et-al-ACM2014, Chen-et-al-IEEE2014, Esser-et-al-2015}.

The most common approach is to compress a trained (full precision) network. HashedNets \citep{chen2015compressing} reduce model sizes by using a hash function to randomly group connection weights and force them to share a single parameter value. \citet{Gong-et-al-2014} compressed deep convnets using vector quantization, which resulteds in only a $1\%$ accuracy loss. However, both methods focused only on the fully connected layers. A recent work by \citet{Han2015} successfully pruned several state-of-the-art large scale networks and showed that the number of parameters could be reduced by an order of magnitude. 

Recent works have shown that more computationally efficient DNNs can
be constructed by quantizing some of the parameters during the training phase. In most cases, DNNs are trained by minimizing some error function using Back-Propagation (BP) or related gradient descent methods. However, such an approach cannot be directly applied if the weights are restricted to binary values. \citet{Soudry-et-al-NIPS2014-small} used a variational Bayesian approach with Mean-Field and Central Limit approximation to calculate the posterior distribution of the weights (the probability of each weight to be +1 or -1). During the inference stage (test phase), their method samples from this distribution one binary network and used it to predict the targets of the test set (More than one binary network can also be used). \citet{Courbariaux2015} similarly used two sets of weights, real-valued and binary. They, however, updated the real valued version of the weights by using gradients computed by applying forward and backward propagation with the set of binary weights (which was obtained by quantizing the real-value weights to +1 and -1).

This study proposes a more advanced technique, referred to as Quantized Neural Network (QNN), for quantizing the neurons and weights during inference and training.  In such networks, all MAC operations can be replaced with $XNOR$ and $population\, count$ (i.e., counting the number of ones in the binary number) operations. This is especially useful in QNNs with the extremely low precision --- for example, when only 1-bit is used per weight and activation, leading to a Binarized Neural Network (BNN). The proposed method is particularly beneficial for implementing large convolutional networks whose neuron-to-weight ratio is very large.\\ 

This paper makes the following contributions:
\begin{itemize}
    
    \item We introduce a method to train Quantized-Neural-Networks (QNNs), neural networks with low precision weights and activations, at run-time, and when computing the parameter gradients at train-time. In the extreme case QNNs use only 1-bit per weight and activation(i.e., Binarized NN; see Section \ref{sec:BNN}).
    
    \item We conduct two sets of experiments, each implemented on a different framework, namely Torch7 and Theano,
        which show that it is possible to train BNNs on MNIST, CIFAR-10 and SVHN and achieve near state-of-the-art results (see Section \ref{sec:benchmark}). Moreover, we report  results on the challenging ImageNet dataset using binary weights/activations as well as quantized version of it (more than 1-bit).
    \item We present preliminary results on quantized gradients and show that it is possible to use only 6-bits with only small accuracy degradation. 
    \item We present results for the Penn Treebank dataset using language models (vanilla RNNs and LSTMs) and show that with 4-bit weights and activations Recurrent QNNs achieve similar accuracies as their 32-bit floating point counterparts. 
    \item  We show that during the forward pass (both at run-time and train-time), QNNs drastically reduce memory consumption (size and number of accesses), and replace most arithmetic operations with bit-wise operations.
        A substantial increase in power efficiency is expected as a result (see Section \ref{sec:power}).
        Moreover, a binarized CNN can lead to binary convolution kernel repetitions; we argue that dedicated hardware could reduce the time complexity by $60\%$ .
    \item Last but not least, we programmed a binary matrix multiplication GPU kernel with which it is possible to run 
        our MNIST BNN 7 times faster than with an unoptimized GPU kernel, 
        without suffering any loss in classification accuracy (see Section \ref{sec:faster}). 
     \item The code for training and applying our BNNs is available on-line (both the Theano \footnote{\url{https://github.com/MatthieuCourbariaux/BinaryNet}}
    and the Torch framework
    \footnote{\url{https://github.com/itayhubara/BinaryNet}}).
\end{itemize}

\section{Binarized Neural Networks}
\label{sec:BNN}
In this section, we detail our binarization function, show how we use it to compute the parameter gradients, and how we backpropagate through it.
\subsection{Deterministic vs Stochastic Binarization}

When training a BNN, we constrain both the weights and the activations to either $+1$ or $-1$.
Those two values are very advantageous from a hardware perspective, as we explain in Section \ref{sec:faster}.
In order to transform the real-valued variables into those two values, 
we use two different binarization functions, as proposed by \cite{Courbariaux-et-al-2015}.
The first binarization function is deterministic:
\begin{equation}
    x^b = {\rm sign}(x) = \left\{ \begin{array}{ll}
                        +1 & \mbox{if $x \geq 0$},\\
                        -1 & \mbox{otherwise},\end{array} \right. 
\end{equation} 
where $x^b$ is the binarized variable (weight or activation) and $x$ the real-valued variable.
It is very straightforward to implement and works quite well in practice.
The second binarization function is stochastic:
\begin{equation}
\label{eq:sampled-wb}
    x^b = \left\{ \begin{array}{ll}
            +1 & \mbox{with probability $p = \sigma(x)$},\\
            -1 & \mbox{with probability $1-p$},\end{array} \right. 
\end{equation}
where $\sigma$ is the {\em ``hard sigmoid''} function:
\begin{equation}
    \sigma(x) = {\rm clip}(\frac{x+1}{2},0,1) = \max(0,\min(1,\frac{x+1}{2})).
\end{equation}
This stochastic binarization is more appealing theoretically (see Section \ref{eq:straight-through-gradient}) than the sign function, 
but somewhat harder to implement as it requires the hardware to generate random bits when quantizing \citep{torii2016asic}.
As a result, we mostly use the deterministic binarization function (i.e., the sign function),
with the exception of {\em activations at train-time} in some of our experiments.

\subsection{Gradient Computation and Accumulation}

Although our BNN training method utilizes binary weights and activations to compute the parameter gradients, the real-valued gradients of the weights are accumulated in real-valued variables, as per Algorithm \ref{alg:train}.
Real-valued weights are likely required for Stochasic Gradient Descent (SGD) to work at all.
SGD explores the space of parameters in small and noisy steps, and that noise is
{\em averaged out} by the stochastic gradient contributions  accumulated in each weight. Therefore, it is important to maintain sufficient resolution for these accumulators, which at first glance
suggests that high precision is absolutely required.

Moreover, adding noise to weights and activations when {\em computing} the parameter gradients provide a form of regularization that can help to generalize better, 
as previously shown with variational weight noise~\citep{Graves-2011-practical},
Dropout~\citep{Srivastava14} and DropConnect~\citep{Wan+al-ICML2013-small}.
Our method of training BNNs can be seen as a variant of Dropout, 
in which instead of randomly setting half of the activations to zero when computing the parameter gradients, we binarize both the activations and the weights.

\subsection{Propagating Gradients Through Discretization}
\label{sec:STE}
The derivative of the sign function is zero almost everywhere, making it apparently incompatible with back-propagation, since the exact gradients of the cost with respect to the quantities before the discretization (pre-activations or weights) are zero. Note that this limitation remains even if stochastic quantization is used.
\citet{Bengio-arxiv2013} studied the question of estimating or propagating gradients through stochastic discrete neurons. He found
that the fastest training was obtained when using
the ``straight-through estimator,'' previously introduced in Hinton's lectures \citep{Hinton-Coursera2012}. We follow a similar approach but use the version of the straight-through
estimator that takes into account the saturation effect, and does use
deterministic rather than stochastic sampling of the bit.
Consider the sign function quantization
\[
q = {\rm Sign}(r),
\]
and assume that an estimator $g_q$ of the gradient $\frac{\partial C}{\partial q}$
has been obtained (with the straight-through estimator when needed).
Then, our straight-through estimator of $\frac{\partial C}{\partial r}$ is simply
\begin{equation}
  \label{eq:straight-through-gradient}
g_r = g_q 1_{|r|\leq 1}.
\end{equation}
Note that this preserves the gradient information and cancels the
gradient when $r$ is too large. 
Not cancelling the gradient when $r$ is too large significantly worsens performance. 
To better understand why the straight-through estimator works well, consider the stochastic binarization scheme in Eq. (\ref{eq:sampled-wb}) and rewrite $\sigma(r)=\left(\mathrm{HT}(r)+1\right)/2$, where $\mathrm{HT}(r)$
is the well-known \textquotedblleft hard tanh\textquotedblright ,
\begin{equation}
    \mathrm{HT}(r) = \left\{ \begin{array}{ll}
            +1 & \mbox{$r>1$},\\
             r & \mbox{$r\in[-1,1]$},\\
            -1 & \mbox{$r<-1$}.\end{array} \right. \label{eq:hard Tanh}
\end{equation}
In this case the input to the next layer has the following form,
\[
\mathbf{W}_{b}h_{b}\left(\mathbf{r}\right)=\mathbf{W}_{b}\mathrm{HT}\left(\mathbf{r}\right)+\mathbf{n}\left(\mathbf{r}\right)\,,\]
where we use the fact that $\mathrm{HT}\left(\mathbf{r}\right)$ is the
expectation over $h_{b}\left(\mathbf{x}\right)$ (see Eqs. (\ref{eq:sampled-wb}) and (\ref{eq:hard Tanh})), and define $\mathbf{n}\left(\mathbf{r}\right)$
as binarization noise with mean equal to zero. When the layer is wide, we expect the deterministic mean term $\mathrm{HT}\left(\mathbf{x}\right)$
to dominate, because the noise term $\mathbf{n}\left(\mathbf{r}\right)$
is a summation over many independent binarizations from all the neurons in
the previous layer. Thus, we argue that the binarization noise $\mathbf{n}\left(\mathbf{x}\right)$ can be ignored when performing differentiation in the backward propagation stage. Therefore, we replace $\frac{\partial h_{b}\left(r\right)}{\partial r}$
(which cannot be computed) with
\begin{equation}
    \frac{\partial\mathrm{HT}\left(\mathbf{r}\right)}{\partial x} 
     = \left\{ \begin{array}{ll}
            0 & \mbox{$r>1$},\\
            1 & \mbox{$r\in[-1,1]$},\\
            0 & \mbox{$r<-1$},\end{array} \right. \label{eq:BackProp}
\end{equation}
which is exactly the straight-through estimator defined in Eq (\ref{eq:straight-through-gradient}).
The use of this straight-through estimator is illustrated in Algorithm \ref{alg:train}.

A similar binarization process was applied for weights in which we combine two ingredients:
\begin{itemize}
\item Project each real-valued weight to [-1,1], i.e., clip the weights
  during training, as per Algorithm \ref{alg:train}. 
  The real-valued weights would otherwise grow very large without any impact on the binary weights.
  \item When using a weight $w^r$, quantize it using $w^b = {\rm Sign}(w^r)$.
\end{itemize}
Projecting the weights to [-1,1] is consistent with the gradient cancelling when $|w^r|>1$, according to Eq. (~\ref{eq:straight-through-gradient}).



\begin{algorithm}
\begin{algorithmic}
    \REQUIRE a minibatch of inputs and targets $(a_0,a^*)$,
    previous weights $W$, previous BatchNorm parameters $\theta$, 
    weight initialization coefficients from \citep{GlorotAISTATS2010-small} $\gamma$,
    and previous learning rate $\eta$.
    \ENSURE updated weights $W^{t+1}$, updated BatchNorm parameters $\theta^{t+1}$ and updated learning rate $\eta^{t+1}$.
    
    \STATE \COMMENT{1. Computing the parameter gradients:}
    
    \STATE \COMMENT{1.1. Forward propagation:}
    \FOR{$k=1$ to $L$}  
        \STATE $W_k^b \leftarrow {\rm Binarize}(W_k)$
        \STATE $s_k \leftarrow a_{k-1}^b W_k^b$
        \STATE $a_k \leftarrow {\rm BatchNorm}(s_k, \theta_k)$
        \IF{$k < L$}
            \STATE $a_k^b \leftarrow {\rm Binarize}(a_k)$
        \ENDIF
    \ENDFOR
    
    \STATE \COMMENT{1.2. Backward propagation:}
    \STATE \COMMENT{Note that the gradients are not binary.}
    \STATE Compute $g_{a_L}=\frac{\partial C}{\partial a_L}$ knowing $a_L$ and $a^*$
    \FOR{$k=L$ to $1$} 
        \IF{$k < L$}
            \STATE $g_{a_k} \leftarrow g_{a_k^b} \circ 1_{|a_k|\leq 1}$
        \ENDIF
        \STATE $(g_{s_k}, g_{\theta_k}) \leftarrow {\rm BackBatchNorm}(g_{a_k}, s_k,\theta_k)$
        
        \STATE $g_{a_{k-1}^b} \leftarrow g_{s_k} W_k^{b}$
        \STATE $g_{W_k^b} \leftarrow g_{s_k}^{\top} a_{k-1}^b$

    \ENDFOR
    
    \STATE \COMMENT{2. Accumulating the parameter gradients:}
    \FOR{$k=1$ to $L$}
        \STATE $\theta_k^{t+1} \leftarrow {\rm Update}(\theta_k, \eta, g_{\theta_k})$
        \STATE $W_k^{t+1} \leftarrow {\rm Clip}({\rm Update}(W_k, \gamma_k \eta, g_{W_k^b}),-1,1)$
        \STATE $\eta^{t+1} \leftarrow \lambda \eta$
    \ENDFOR
    
\end{algorithmic}
\caption{
Training a BNN. $C$ is the cost function for minibatch, $\lambda$, the learning rate decay factor, and $L$, the number of layers. $(\circ)$ stands for element-wise multiplication. The function Binarize($\cdot$) specifies how to (stochastically or deterministically) binarize the activations and weights, and Clip(), how to clip the weights. BatchNorm() specifies how to batch-normalize the activations, using either batch normalization \citep{Ioffe+Szegedy-2015} or its shift-based variant we describe in Algorithm \ref{alg:BN}.
BackBatchNorm() specifies how to backpropagate through the normalization.
Update() specifies how to update the parameters when their gradients are known,
using either ADAM \citep{kingma2014adam} or the shift-based AdaMax we describe in Algorithm \ref{alg:adamax}.
}
\label{alg:train}
\end{algorithm}


\begin{algorithm} [ht]
\begin{algorithmic}
    \REQUIRE Values of $x$ over a mini-batch: $B=\{x_{1\ldots m}\}$; 
        Parameters to be learned: $\gamma$, $\beta$
    \ENSURE $\{y_i = {\rm BN}({x_i,}{\gamma,\beta})\}$
    \STATE $\mu_B \leftarrow \frac{1}{m}\sum_{i=1}^m x_i$ \COMMENT{mini-batch mean}
    \STATE $C(x_i) \leftarrow (x_i-\mu_B)$
    \COMMENT{centered input}
    \STATE $\sigma_B^2\! \leftarrow \!\! \frac{1}{m}\sum_{i=1}^m\! (C(x_i)\!\!\ll\gg\!\! \mathrm{AP2}(C(x_i)))\!$ \COMMENT{apx variance}
    \STATE $\hat{x_i} \leftarrow C(x_i)\ll\gg \mathrm{AP2}((\sqrt{\sigma_B^2+\epsilon})^{-1})$ \COMMENT{normalize}
    \STATE $y_i \leftarrow \mathrm{AP2}(\gamma)\ll\gg\hat{x_i}$ \COMMENT{scale and shift} 
\end{algorithmic}
\caption{Shift based Batch Normalizing Transform, applied to activation $x$ over a mini-batch. $\mathrm{AP2}(x) = \mathrm{sign}(x) \times 2^{\mathrm{round}(\mathrm{log2}{|x|})}$
    is the approximate power-of-2 \protect\footnotemark, and  $\ll\gg$ stands for \textbf{both }left and right binary shift.}
\label{alg:BN}
\end{algorithm}

\footnotetext{Hardware implementation of $\mathrm{AP2}$ is as simple as extracting the index of the most significant bit from the number's binary representation.}


\begin{algorithm}[ht]
\begin{algorithmic}
    \REQUIRE Previous parameters $\theta_{t-1}$, their gradient $g_t$, and learning rate $\alpha$.
    \ENSURE Updated parameters $\theta_t$
    \STATE \COMMENT{Biased 1st and 2nd raw moment estimates:}
    \STATE $m_t \gets \beta_1 \cdot m_{t-1} + (1-\beta_1) \cdot g_t$  
    \STATE $v_t \gets \max (\beta_2 \cdot v_{t-1} , |g_t| )$ 
    \STATE \COMMENT{Updated parameters:}
    \STATE $\theta_t \gets \theta_{t-1} - (\alpha \ll\gg (1-\beta_1)) \cdot \hat{m}\ll\gg v_t^{-1})$
\end{algorithmic}
\caption{Shift based AdaMax learning rule \citep{kingma2014adam}.
$g^2_t$ indicates the element-wise square $g_t \circ g_t$. 
Good default settings are $\alpha=2^{-10}$, $1-\beta_1=2^{-3}$, $1-\beta_2=2^{-10}$. 
All operations on vectors are element-wise. 
With $\beta_1^t$ and $\beta_2^t$ we denote $\beta_1$ and $\beta_2$ to the power $t$.}
\label{alg:adamax}
\end{algorithm}
\begin{algorithm}[ht]
\begin{algorithmic}
    \REQUIRE 8-bit input vector $a_0$, 
     binary weights $W^b$, and BatchNorm parameters $\theta$.
    \ENSURE the MLP output $a_L$.
    
    \STATE \COMMENT{1. First layer:}
    \STATE $a_1 \leftarrow 0$
    \FOR{$n=1$ to $8$}  
        \STATE $a_1 \leftarrow a_1 + 2^{n-1} \times {\rm XnorDotProduct(a_0^n,W^b_1)}$
    \ENDFOR
    \STATE $a_1^b \leftarrow {\rm Sign(BatchNorm}(a_1,\theta_1))$
    
    \STATE \COMMENT{2. Remaining hidden layers:}
    \FOR{$k=2$ to $L-1$}  
        \STATE $a_k \leftarrow {\rm XnorDotProduct}(a_{k-1}^b,W^b_k)$
        \STATE $a_k^b \leftarrow {\rm Sign(BatchNorm}(a_k,\theta_k))$
    \ENDFOR
    
    \STATE \COMMENT{3. Output layer:}
    \STATE $a_L \leftarrow {\rm XnorDotProduct}(a_{L-1}^b,W^b_L)$
    \STATE $a_L \leftarrow {\rm BatchNorm}(a_L,\theta_L)$
    
\end{algorithmic}
\caption{Running a BNN with $L$ layers.}
\label{alg:run}
\end{algorithm}

\subsection{Shift-based Batch Normalization} 
\label{sec:SBN}
Batch Normalization (BN) \citep{Ioffe+Szegedy-2015} accelerates the training and reduces the overall impact of the weight scale \citep{Courbariaux-et-al-2015}. 
The normalization procedure may also help to regularize the model. 
However, at train-time, BN requires many multiplications (calculating the standard deviation and dividing by it,
namely, dividing by the running variance, which is the weighted mean of the training set activation variance). 
Although the number of scaling calculations is the same as the number of neurons, in the case of ConvNets this number is quite large. 
For example, in the CIFAR-10 dataset (using our architecture), the first convolution layer, consisting of only $128\times3\times3$ filter masks, 
converts an image of size $3\times32\times32$ to size $128\times28\times28$, 
which is almost two orders of magnitude larger than the number of weights (87.1 to be exact). To achieve the results that BN would obtain, 
we use a shift-based batch normalization (SBN) technique, presented in Algorithm \ref{alg:BN}.
SBN approximates BN almost without multiplications. Define $\mathrm{AP2}(z)$ as the approximate power-of-2 of z (i.e., the index of the most significant bit (MSB)), and $\ll\gg$ as both left and right binary shift. SBN replaces almost all multiplication with power-of-2-approximation and shift operations:
\begin{equation}
x\times y\rightarrow x\ll\gg \mathrm{AP2}(y).
\end{equation}

The only operation which is not a binary shift or an add is the inverse square root (see normalization operation Algorithm \ref{alg:BN}). From the early work of \citet{lomont2003fast} we know that the inverse-square operation could be applied with approximately the same complexity as multiplication. There are also faster methods, which involve lookup table tricks that typically obtain lower accuracy (this may not be an issue, since our procedure already adds a lot of noise). However, the number of values on which we apply the inverse-square operation is rather small, since it is done after calculating the variance, i.e., after averaging  (for a more precise calculation, see the BN analysis in \citet{Lin2015}.  Furthermore, the size of the standard deviation vectors is relatively small. For example, these values make up only $0.3\%$ of the network size (i.e., the number of learnable parameters) in the Cifar-10 network we used in our experiments. 

In the experiment we observed no loss in accuracy when using the shift-based BN algorithm instead of the vanilla BN algorithm.
 
\subsection{Shift Based AdaMax} 
 
The ADAM learning method \citep{kingma2014adam} also reduces the impact of the weight scale. 
Since ADAM requires many multiplications, 
we suggest using instead the shift-based AdaMax we outlined in Algorithm \ref{alg:adamax}. 
In the experiment we conducted we observed no loss in accuracy when using the shift-based AdaMax algorithm instead of the vanilla ADAM algorithm.

 

\subsection{First Layer}

In a BNN, only the binarized values of the weights and activations are used in all calculations.
As the output of one layer is the input of the next, the inputs of all the layers are binary, with the exception of the first layer.
However, we do not believe this to be a major issue.
First, in computer vision, the input representation typically has far fewer channels (e.g, red, green and blue) 
than internal representations (e.g., 512).
Consequently, the first layer of a ConvNet is often the smallest convolution layer, 
both in terms of parameters and computations \citep{Szegedy-et-al-arxiv2014}. Second, it is relatively easy to handle continuous-valued inputs as fixed point numbers,
with $m$ bits of precision. For example, in the common case of $8$-bit fixed point inputs:
\begin{equation}
\label{eq:FirstLayer}
     s = x \cdot w^b, \qquad \qquad  \qquad\qquad     s = \sum_{n=1}^{8} {2^{n-1} (x^n \cdot w^b),}
\end{equation}
where $x$ is a vector of 1024 8-bit inputs, $x_1^8$ is the most significant bit of the first input, $w^b$ is a vector of 1024 1-bit weights,
and $s$ is the resulting weighted sum.
This method is used in Algorithm \ref{alg:run}.

\section{Qunatized Neural network - More than 1-bit}
\label{sec:QNN}
Observing Eq.~(\ref{eq:FirstLayer}), we can see that using 2-bit activations simply doubles the number of times we need to run our XnorPopCount Kernel (i.e., directly proportional to the activation bitwidth). This idea was recently proposed by \cite{zhou2016dorefa} (DoReFa net) and \cite{miyashita2016convolutional} (published on arXive shortly after our preliminary technical report was published there). However, in contrast to to \citeauthor{zhou2016dorefa}, we did not find it useful to initialize the network with weights obtained by training the network with full precision weights. Moreover, the \citeauthor{zhou2016dorefa} network did not quantize the weights of the first convolutional layer and the last fully-connected layer, whereas we binarized both.  We followed the quantization schemes suggested by \cite{miyashita2016convolutional}, namely, linear quantization:
\begin{equation}
\mathbf{LinearQuant}(x, bitwidth)=\mathrm{Clip}\left(\mathrm{round}\left(\frac{x}{bitwidth}\right)\times bitwidth,minV,maxV\right)
\end{equation}
and logarithmic quantization:
\begin{equation}
\begin{aligned}
&\mathbf{LogQuant}(x, bitwidth)\left(\mathbf{x}\right) =\mathrm{Clip}\left(\mathrm{AP2}(x),minV,maxV\right),
\end{aligned}
\end{equation}
where $minV$ and $maxV$ are the minimum and maximum scale range respectively. Where $\mathrm{AP2}(x)$ is the approximate-power-of-2 of $x$ as described in Section  \ref{sec:SBN}. In our experiments (detailed in Section \ref{sec:benchmark}) we applied the above quantization schemes on the weights, activations and gradients and tested them on the more challenging ImageNet dataset.

\section{Benchmark Results}
\label{sec:benchmark}

\subsection{Results on MNIST,SVHN, and CIFAR-10}
\begin{table*}[ht]
\protect\caption{Classification test error rates of DNNs trained on MNIST (fully connected architecture), CIFAR-10 and SVHN (convnet). No unsupervised pre-training or data augmentation was used.}
\centering{}%

\begin{tabular}{lcccr}
\hline 
Data set  & MNIST & SVHN & CIFAR-10\tabularnewline
\hline 
\hline 
\multicolumn{5}{c}{Binarized activations+weights, during training and test}\tabularnewline
\hline 
\hline 
BNN (Torch7) & 1.40\% & 2.53\% & 10.15\% & \tabularnewline
BNN (Theano) & 0.96\% & 2.80\% & 11.40\% & \tabularnewline
Committee Machines' Array \cite{Baldassi2015} & 1.35\% & - & - \tabularnewline
\hline 
\hline 
\multicolumn{5}{c}{Binarized weights, during training and test}\tabularnewline
\hline 
\hline 
BinaryConnect \cite{Courbariaux-et-al-2015} & 1.29$\pm$ 0.08\% & 2.30\% & 9.90\% & \tabularnewline
\hline 
\hline 
\multicolumn{5}{c}{Binarized activations+weights, during test}\tabularnewline
\hline 
\hline 
EBP \cite{Cheng-et-al-2015} & 2.2$\pm$ 0.1\% & - & - & \tabularnewline
Bitwise DNNs \cite{Kim-et-al-2016} & 1.33\% & - & - & \tabularnewline
\hline 
\hline 
\multicolumn{5}{c}{Ternary weights, binary activations, during test}\tabularnewline
\hline 
\hline 
\cite{hwang-et-al-2014} & 1.45\% & - & - & \tabularnewline

\hline 
\hline 
\multicolumn{5}{c}{No binarization (standard results)}\tabularnewline
\hline 
\hline 
No reg & 1.3$\pm$ 0.2\% & 2.44\% & 10.94\% & \tabularnewline
Maxout Networks \cite{Goodfellow2013a} & 0.94\% & 2.47\% & 11.68\% & \tabularnewline
Gated pooling \cite{lee-et-al-2015}& - & 1.69\% & 7.62\% & \tabularnewline
\hline 
\end{tabular}\label{tab:benchmarks}

\end{table*}

\begin{figure}
\caption{Training curves for different methods on the CIFAR-10 dataset.
   The dotted lines represent the training costs (square hinge losses)
   and the continuous lines the corresponding validation error rates.
   Although BNNs are slower to train, they are nearly as accurate as 32-bit float DNNs.}
  \begin{centering}
    \includegraphics[scale=0.5]{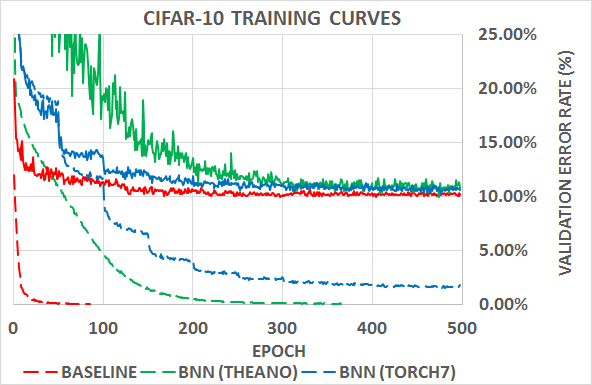}
  \par\end{centering}
\label{fig:curves}
\end{figure}

We performed two sets of experiments, each based on a different framework, namely Torch7 and Theano. 
Other than the framework, the two sets of experiments are very similar:
\begin{itemize}
    \item In both sets of experiments, we obtain near state-of-the-art results with BNNs on MNIST, CIFAR-10 and the SVHN benchmark datasets.
    \item In our Torch7 experiments, the activations are {\em stochastically} binarized at train-time,
        whereas in our Theano experiments they are {\em deterministically} binarized.
    \item In our Torch7 experiments, we use the {\em shift-based BN and AdaMax} variants, 
        which are detailed in Algorithms \ref{alg:BN} and \ref{alg:adamax},
        whereas in our Theano experiments, we use {\em vanilla BN and ADAM}.
\end{itemize} 

Results are reported in Table \ref{tab:benchmarks}. Implementation details are reported in Appendix A.
\paragraph{MNIST} MNIST is an image classification benchmark dataset \citep{LeCun+98}.
It consists of a training set of 60K and a test set of 10K 28 $\times$ 28 gray-scale images representing digits ranging from 0 to 9.
The Multi-Layer-Perceptron (MLP) we train on MNIST consists of 3 hidden layers. In our Theano implementation we used hidden layers of size 4096 whereas in our Torch implementation we used much smaller size 2048. This difference explains the accuracy gap between the two implementations. 
\paragraph{CIFAR-10} CIFAR-10 is an image classification benchmark dataset.
It consists of a training set of size 50K and a test set of size 10K, where instances are 32 $\times$ 32 color images representing airplanes, automobiles, birds, cats, deer, dogs, frogs, horses, ships and trucks. Both implementations share the same structure as reported in Appendix A. Since the Torch implementation uses stochastic binarization, it achieved slightly better results. 
\paragraph{SVHN} Street View House Numbers (SVHN) is also an image classification benchmark dataset.
It consists of a training set of size 604K examples and a test set of size 26K, where instances are 32 $\times$ 32 color images representing digits ranging from 0 to 9. Here again we obtained a small improvement in the performance by using stochastic binarization scheeme. 

\subsection{Results on ImageNet} 
To test the strength of our method, we applied it to the challenging ImageNet classification task, which is probably the most important classification benchmark dataset.
It consists of a training set of size 1.2M samples and test set of size 50K. Each instance is labeled with one of 1000 categories including objects, animals, scenes, and even some abstract shapes. On ImageNet, it is customary to report two error rates: top-1 and top-5, where the top-$x$ error rate is the fraction of test images for which the correct label is not among the $x$ labels considered most probable by the model. Considerable research has been concerned with compressing ImageNet architectures while preserving high accuracy. Previous approaches include pruning near zero weights \citep{Gong-et-al-2014, han2015deep} using matrix factorization techniques \citep{zhang2015efficient}, quantizing the weights \citep{gupta2015deep}, using shared weights \citep{chen2015compressing} and applying Huffman codes \citep{han2015deep} among others.

To the best of our knowledge, before the first revision of this paper was published on arXive, no one had reported on successfully quantizing the network's activations. On the contrary, a recent work \citep{han2015deep} showed that accuracy significantly deteriorates when trying to quantize convolutional layers' weights below 4-bit (FC layers are more robust to quantization and can operate quite well with only 2 bits). 
In the present work we attempted to tackle the difficult task of binarizing 
both weights and activations. Employing the well-known AlexNet and GoogleNet architectures, we applied our techniques and achieved $41.8\%$ top-1 and $67.1\%$ top-5 accuracy using AlexNet and $47.1\%$ top-1 and $69.1\%$ top-5 accuracy using GoogleNet. While these performance results leave room for improvement (relative to full precision nets), they are by far better than all previous attempts to compress ImageNet architectures using less than 4-bit precision for the weights. Moreover, this advantage is achieved while also binarizing neuron activations.

\subsection{Relaxing \textquotedblleft hard tanh\textquotedblright $ $  boundaries}
We discovered that after training the network it is useful to widen the \textquotedblleft hard tanh\textquotedblright $ $  boundaries and retrain the network. As explained in Section \ref{sec:STE}, the straight-through estimator (which can be written as \textquotedblleft hard tanh\textquotedblright) cancels gradients coming from neurons with absolute values higher than 1. Hence, towards the last training iterations most of the gradient values are zero and the weight values cease to update. By relaxing the \textquotedblleft hard tanh\textquotedblright $ $ boundaries we allow more gradients to flow in the back-propagation phase and improve top-1 accuracies by $1.5\%$ on AlexNet topology using vanilla BNN.

\subsection{2-bit activations} 
While training BNNs on the ImageNet dataset we noticed that we could not force the training set error rate to converge to zero. In fact the training error rate stayed fairly close to the validation error rate. This observation led us to investigate a more relaxed activation quantization (more than 1-bit).
As can be seen in Table \ref{tab:AlexNet}, the results are quite impressive and illustrate an approximate $5.6\%$ drop in performance (top-1 accuracy) relative to floating point representation, using only 1-bit weights and 2-bit activation. Following \cite{miyashita2016convolutional}, we also tried quantizing the gradients and discovered that only logarithmic quantization works. With 6-bit gradients we achieved $46.8\%$ degradation. Those results are presently state-of-the-art, surpassing those obtained by the DoReFa net \citep{zhou2016dorefa}. As opposed to DoReFa,  we utilized a deterministic quantization process rather than a stochastic one. Moreover, it is important to note that while quantizing the gradients, DoReFa assigns for each instance in a mini-batch its own scaling factor, which increases the number of MAC operations.

While AlexNet can be compressed rather easily, compressing GoogleNet is much harder due to its small number of parameters. When using vanilla BNNs, we observed a large degradation in the top-1 results. However, by using QNNs with 4-bit weights and activation, we were able to achieve $66.5\%$ top-1 accuracy (only a $5.5\%$ drop in performance compared to the 32-bit floating point architecture), which is the current state-of-the-art-compression result over GoogleNet. Moreover, by using QNNs with 6-bit weights, activations and gradients we achieved $66.4\%$ top-1 accuracy. Full implementation details of our experiments are reported in Appendix \ref{appendix-ImageNet}.
\begin{table*}[ht]
\protect\caption{Classification test error rates of the AlexNet model trained on the ImageNet 1000 classification task. No unsupervised pre-training or data augmentation was used.}
\centering{}%
\scalebox{1}{
\begin{tabular}{lcccr}
\hline 
Model & Top-1 & Top-5 &&\tabularnewline
\hline 
\hline 
\multicolumn{5}{c}{Binarized activations+weights, during training and test}\tabularnewline
\hline 
\hline 
BNN  & 41.8\% & 67.1\% & \tabularnewline
Xnor-Nets\tablefootnote{\label{first_last} First and last layers were not binarized (i.e., using 32-bit precision weights and activation)} ~\citep{rastegari2016xnor} & 44.2\% & 69.2\%   \tabularnewline
\hline 
\hline 
\multicolumn{5}{c}{Binary weights and Quantize activations during training and test}\tabularnewline
QNN 2-bit activation  & 51.03\% & 73.67\% &  \tabularnewline
DoReFaNet 2-bit activation\footref{first_last} \citep{zhou2016dorefa} & 50.7\% & 72.57\% & \tabularnewline
\hline 
\hline 
\multicolumn{5}{c}{Quantize weights, during test}\tabularnewline
\hline 
\hline 
Deep Compression 4/2-bit (conv/FC layer) \citep{han2015deep} & 55.34\% & 77.67\% &  \tabularnewline
\citep{gysel2016hardware} - 2-bit  & 0.01\% & -\% &  \tabularnewline
\hline 
\hline 
\multicolumn{5}{c}{No Quantization (standard results)}\tabularnewline
\hline 
\hline 
AlexNet - our implementation & 56.6\% & 80.2\% &  \tabularnewline
\hline 
\end{tabular}\label{tab:AlexNet}
}
\end{table*}

\begin{table*}[h]
\protect\caption{Classification test error rates of the GoogleNet model trained on the ImageNet 1000 classification task. No unsupervised pre-training or data augmentation was used.}
\centering{}%
\scalebox{1}{
\begin{tabular}{lcccr}
\hline 
Model  & Top-1 & Top-5 &&\tabularnewline
\hline 
\hline 
\multicolumn{5}{c}{Binarized activations+weights, during training and test}\tabularnewline
\hline 
\hline 
BNN  & 47.1\% & 69.1\% & \tabularnewline
\hline
\multicolumn{5}{c}{Quantize weights and activations during training and test}\tabularnewline 
\hline 
\hline 
QNN 4-bit & 66.5\% & 83.4\% & \tabularnewline
\hline 
\hline 
\multicolumn{5}{c}{Quantize activation,weights and gradients during training and test}\tabularnewline 
\hline 
\hline 
QNN 6-bit & 66.4\% & 83.1\% &
\tabularnewline
\hline
\hline 
\multicolumn{5}{c}{No Quantization (standard results)}\tabularnewline
\hline 
\hline 
GoogleNet - our implementation & 71.6\% & 91.2\% &  \tabularnewline
\hline 
\end{tabular}\label{tab:GoogleNet}
}
\end{table*}

\label{sec:preliminary}
 
\subsection{Language Models}
Recurrent neural networks (RNNs) are very demanding in memory and computational power in comparison to feed forward networks. There are a large variety of recurrent models with the Long Short Term Memory networks (LSTMs) introduced by \citet{hochreiter1997long} are being the most popular model. LSTMs are a special kind of RNN, capable of learning long-term dependencies using unique gating mechanisms. Recently, \citet{ott2016recurrent} tried to quantize the RNNs weight matrices using similar techniques as described in Section \ref{sec:BNN}.  They observed that the weight binarization methods do not work with RNNs. However, by using 2-bits (i.e., ${-1,0,1}$), they have been able to achieve similar and even higher accuracy on several datasets. Here we report on the first attempt to quantize both weights and activations by trying to evaluate the accuracy of quantized recurrent models  trained on the Penn Treebank dataset. The Penn Treebank Corpus \citep{marcus1993building} contains 10K unique words. We followed the same setting as in \citep{mikolov2012context} which resulted in 18.55K words for training set, 14.5K and 16K words in the validation and test sets respectively. We experimented with both vanilla RNNs  and LSTMs. For our vanilla RNN model we used one hidden layers of size 2048 and ReLU as the activation function. For our LSTM model we use 1 hidden layer of size 300. Our RNN implementation was constructed to predict the next character hence performance was measured using the bits-per-character (BPC) metric. In the LSTM model we tried to predict the next word so performance was measured using the perplexity per word (PPW) metric. Similar to  \citep{ott2016recurrent}, our preliminary results indicate that binarization of weight matrices lead to large accuracy degradation. However, as can be seen in Table \ref{tab:LangModels}, with 4-bits activations and weights we can achieve similar accuracies as their 32-bit floating point counterparts. 

\begin{table*}[h]
\protect\caption{Language Models results on Penn Treebank dataset.}{Language Models results on Penn Treebank dataset. FP stands for 32-bit floating point}
\centering{}%
\scalebox{1}{
\begin{tabular}{lccccccr}
\hline 
Model  & Layers & Hidden Units & bits(weights) & bits(activation) & Accuracy &&\tabularnewline
\hline 
\hline 
RNN    & 1 & 2048 & 3  & 3  &1.81 BPC &  \tabularnewline
RNN    & 1 & 2048 & 2  & 4  &1.67  BPC &  \tabularnewline
RNN    & 1 & 2048 & 3  & 4  &1.11 BPC &  \tabularnewline
RNN    & 1 & 2048 & 3  & 4  &1.05 BPC & \tabularnewline
RNN    & 1 & 2048 & FP & FP &1.05 BPC & \tabularnewline
\hline 
\hline 
LSTM    & 1 & 300 & 2  & 3  & 220 PPW &  \tabularnewline
LSTM    & 1 & 300 & 3  & 4  & 110 PPW&  \tabularnewline
LSTM    & 1 & 300 & 4  & 4  & 100 PPW&  \tabularnewline
LSTM    & 1 & 900 & 4  & 4  & 97 PPW&  \tabularnewline
LSTM    & 1 & 300 & FP & FP & 97 PPW&  \tabularnewline
\end{tabular}\label{tab:LangModels}
}
\end{table*}

\section{High Power Efficiency during the Forward Pass}
\label{sec:power}

\begin{table}[!htb]
      \caption{Energy consumption of multiply- accumulations; see  \cite{Horowitz2014}}
      \centering
        \begin{tabular}{llc}
	        \hline 
			Operation & MUL  & ADD \tabularnewline
            \hline 
             8-bit Integer & 0.2pJ & 0.03pJ\tabularnewline
            32-bit Integer  & 3.1pJ & 0.1pJ\tabularnewline
            16-bit Floating Point & 1.1pJ & 0.4pJ\tabularnewline
            32-bit Floating Point  & 3.7pJ & 0.9pJ\tabularnewline
            \hline 
        \end{tabular}
        \label{TB:Memory}
      \centering
        \caption{Energy consumption of memory accesses; see \cite{Horowitz2014} }
        \begin{tabular}{llc}
	        \hline 
			Memory size  & 64-bit Cache \tabularnewline
            \hline 
            8K & 10pJ\tabularnewline
            32K & 20pJ\tabularnewline
            1M & 100pJ\tabularnewline
            DRAM & 1.3-2.6nJ\tabularnewline
            \hline 
        \end{tabular}
        \label{Tb:Add_MUL_Horowitz}
\end{table}

Computer hardware, be it general-purpose or specialized, is composed of memories, arithmetic operators and control logic.
During the forward pass (both at run-time and train-time), BNNs drastically reduce memory size and accesses, 
and replace most arithmetic operations with bit-wise operations,
which might lead to vastly improved power-efficiency.
Moreover, a binarized CNN can lead to binary convolution kernel repetitions, and we argue that dedicated hardware could reduce the time complexity by $60\%$ .
\begin{figure}[ht]
\caption{Binary weight filters, sampled from of the first convolution layer.
Since we have only $2^{k^{2}}$ unique 2D filters (where $k$ is the
filter size), filter replication is very common.
For instance, on our CIFAR-10 ConvNet,
only 42\% of the filters are unique. }
\begin{centering}
\includegraphics[viewport=140bp 220bp 480bp 550bp,clip,width=0.5\columnwidth]{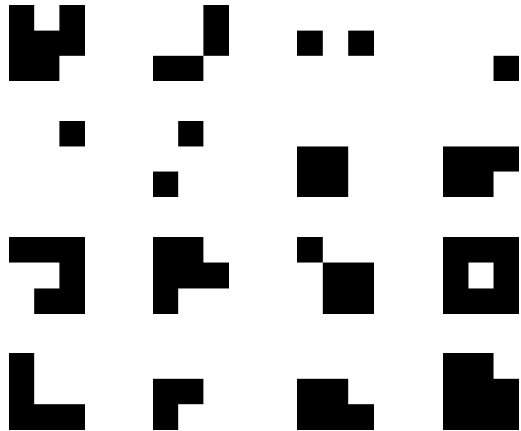}
\par\end{centering}
\label{Fig:filters}
\end{figure}

\paragraph{Memory Size and Accesses}

Improving computing performance has always been and remains a challenge.
Over the last decade, power has been the main constraint on performance \citep{Horowitz2014}.
This is why considerable research efforts have been devoted to reducing the energy consumption of neural networks. 
\citet{Horowitz2014} provides rough numbers for the energy consumed by the computation (the given numbers are for 45nm technology),
as summarized in Tables \ref{TB:Memory} and \ref{Tb:Add_MUL_Horowitz}.
Importantly, we can see that memory accesses typically consume more energy than arithmetic operations,
and {\em memory access cost increases with memory size}.
In comparison with 32-bit DNNs, BNNs require 32 times smaller memory size {\em and} 32 times fewer memory accesses. This is expected to reduce energy consumption drastically (i.e., by a factor larger than 32).

\paragraph{XNOR-Count}

Applying a DNN mainly involves convolutions and matrix multiplications.
The key arithmetic operation of deep learning is thus the multiply-accumulate operation. 
Artificial neurons are basically multiply-accumulators computing weighted sums of their inputs.
In BNNs, both the activations and the weights are constrained to either $-1$ or $+1$.
As a result, most of the 32-bit floating point multiply-accumulations are replaced by 1-bit XNOR-count operations.
This could have a big impact on dedicated deep learning hardware.
For instance, a 32-bit floating point multiplier costs about 200 Xilinx FPGA slices \citep{Govindu-et-al-2004,Beauchamp-et-al-2006},
whereas a 1-bit XNOR gate only costs a single slice.

When using a ConvNet architecture with binary weights, the number of unique
filters is bounded by the filter size. For example, in our implementation
we use filters of size $3\times3$, so the maximum number of unique
2D filters is $2^{9}=512$. However, this should not prevent expanding
the number of feature maps beyond this number, since the actual filter
is a 3D matrix. Assuming we have $M_{\ell}$ filters in the $\ell$
convolutional layer, we have to store a 4D weight matrix of size $M_{\ell}\times M_{\ell-1}\times k\times k$.
Consequently, the number of unique filters is $2^{k^{2}M_{\ell-1}}$.
When necessary, we apply each filter on the map and perform the required multiply-accumulate (MAC) operations (in our case, using XNOR and popcount operations).
Since we now have binary filters, many 2D filters of size $k\times k$
repeat themselves. By using dedicated hardware/software, we can apply
only the unique 2D filters on each feature map and sum the results to receive each 3D filter's convolutional result. Note that an
inverse filter (i.e., {[}-1,1,-1{]} is the inverse of {[}1,-1,1{]})
can also be treated as a repetition; it is merely a multiplication
of the original filter by -1. For example, in our ConvNet architecture
trained on the CIFAR-10 benchmark, there are only 42\% unique filters
per layer on average. Hence we can reduce the number of the XNOR-popcount operations by 3. 

QNNs complexity scale up linearly with the number of bits per weight/activation, since it requires the application of the XNOR kernel several times (see Section \ref{sec:QNN}). As of now, QNNs still supply the best compression to accuracy ratio. Moreover, quantizing the gradients allows us to use the XNOR kernel for the backward pass,  leading to fully fixed point layers with low bitwidth. By accelerating the training phase, QNNs can play an important role in future power demanding tasks. 

\section{Seven Times Faster on GPU at Run-Time}
\label{sec:faster}

It is possible to speed up GPU implementations of QNNs,
by using a method sometimes called SIMD (single instruction, multiple data) within a register (SWAR). 
The basic idea of SWAR is to {\em concatenate} groups of 32 binary variables into 32-bit registers,
and thus obtain a 32-times speed-up on bitwise operations (e.g., XNOR).
Using SWAR, it is possible to evaluate 32 connections with only 3 instructions:
\begin{align}
    \label{eq:popc}
    a_1 += {\rm popcount(xnor}(a_0^{32b},w^{32b}_1)),
\end{align}
where $a_1$ is the resulting weighted sum, and $a_0^{32b}$ and $w^{32b}_1$ are the concatenated inputs and weights.
Those 3 instructions (accumulation, popcount, xnor) take $1+4+1=6$ clock cycles on recent Nvidia GPUs 
(and if they were to become a fused instruction, it would only take a single clock cycle).
Consequently, we obtain a theoretical Nvidia GPU speed-up of factor of $32/6 \approx 5.3$.
In practice, this speed-up is quite easy to obtain
as the memory bandwidth to computation ratio is also increased 6 times.

\begin{figure}[h]
\caption{
The first 3 columns show the time it takes to perform a $8192\times8192\times8192$ (binary) matrix multiplication on a GTX750 Nvidia GPU, depending on which kernel is used. The next three columns show the time it takes to run the MLP from Section 3 on the full MNIST test set. The last three columns show that the MLP accuracy does not depend on the kernel
}
\begin{center}
\centerline{\includegraphics[width=.5\textwidth]{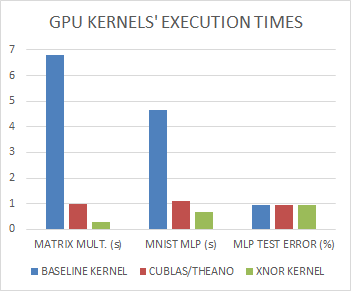}}
\end{center}
\label{fig:kernels}

\end{figure}

In order to validate those theoretical results, we programmed two GPU kernels:
\begin{itemize}
    \item An unoptimized matrix multiplication kernel that serves as our baseline. 
    \item The XNOR kernel, which is nearly identical to the baseline, 
        except that it uses the SWAR method, as in Equation (\ref{eq:popc}).
\end{itemize}
The two GPU kernels return identical outputs when their inputs are constrained to $-1$ or $+1$ (but not otherwise).
The XNOR kernel is about {\em 23 times faster than the baseline kernel} and {\em 3.4 times faster than cuBLAS}, 
as shown in Figure \ref{fig:kernels}.
Last but not least, the MLP from Section \ref{sec:benchmark} runs 7 times faster with the XNOR kernel
than with the baseline kernel, without suffering any loss in classification accuracy (see Figure \ref{fig:kernels}). As MNIST's images are not binary, the first layer's computations are always performed by the baseline kernel. The last three columns show that the MLP accuracy does not depend on which kernel is used.

\section{Discussion and Related Work} 

Until recently, the use of extremely low-precision networks (binary in the extreme case) was believed to substantially degrade the network performance \citep{courbariaux+al-TR2014}. \citet{Soudry-et-al-NIPS2014-small} and \citet{Cheng-et-al-2015} proved the contrary by showing that good performance could be achieved even if all neurons and weights are binarized to $\pm 1$ . This was done using Expectation BackPropagation (EBP), a variational Bayesian approach, which infers networks with binary weights and neurons by updating the posterior distributions over the weights. These distributions are updated by differentiating their parameters (e.g., mean values) via the back propagation (BP) algorithm. \citet{Esser-et-al-2015} implemented a fully binary network at run time using a very similar approach to EBP, showing significant improvement in energy efficiency. The drawback of EBP is that the binarized parameters are only used during inference. 

The probabilistic idea behind EBP was extended in the BinaryConnect algorithm of \citet{Courbariaux-et-al-2015}. In BinaryConnect, the real-valued version of the weights is saved and used as a key reference for the binarization process. The binarization noise is independent between different weights, either by construction (by using stochastic quantization) or by assumption (a common simplification; see  \citealp{spang1962reduction}). The noise would have little effect on the next neuron's input because the input is a summation over many weighted neurons. Thus, the real-valued version could be updated using the back propagated error by simply ignoring the binarization noise in the update. With this method, \citet{Courbariaux-et-al-2015} were the first to binarize weights in CNNs and achieved near state-of-the-art performance on several datasets. They also argued that noisy weights provide a form of regularization, which could help to improve generalization, as previously shown by \cite{Wan+al-ICML2013-small}. This method binarized weights while still maintaining full precision neurons.

\citet{Lin-et-al-2015} carried over the work of \citet{Courbariaux-et-al-2015} to the back-propagation process by quantizing the representations at each layer of the network, to convert some of the remaining multiplications into binary shifts by restricting the neurons' values to be power-of-two integers. \citet{Lin-et-al-2015}'s work and ours seem to share similar characteristics .However, their approach continues to use full precision weights during the test phase. Moreover, \citet{Lin-et-al-2015} quantize the neurons only during the back propagation process, and not during forward propagation. 

Other research \citep{Baldassi2015} showed that full binary training and testing is possible in an array of committee machines with randomized input, where only one weight layer is being adjusted. \citet{Gong-et-al-2014} aimed to compress a fully trained high precision network by using quantization or matrix factorization methods. These methods required training the network with full precision weights and neurons, thus requiring numerous MAC operations (which the proposed QNN algorithm avoids). \citet{hwang-et-al-2014} focused on a fixed-point neural network design and achieved performance almost identical to that of the floating-point architecture. \citet{Kim-et-al-2016} \emph{retrained} neural networks with binary weights and activations.

As far as we know, before the first revision of this paper was published on arXive, no work succeeded in binarizing weights \emph{and} neurons, at the inference phase \emph{and} the entire training phase of a deep network. This was achieved in the present work. We relied on the idea that binarization can be done stochastically, or be approximated as random noise. This was previously done for the weights by \citet{Courbariaux-et-al-2015}, but our BNNs extend this to the activations. Note that the binary activations are especially important for ConvNets, where there are typically many more neurons than free weights. This allows highly efficient operation of the binarized DNN at run time, and at the forward-propagation phase during training. Moreover, our training method has almost no multiplications, and therefore might be implemented efficiently in dedicated hardware. However, we have to save the value of the full precision weights. This is a remaining computational bottleneck during training, since it is an energy-consuming operation.  

Shortly after the first version of this paper was posted on arXiv, several papers tried to improve and extend it. \citet{rastegari2016xnor} made a small modification to our algorithm (namely multiplying the binary weights and input by their $L_1$ norm) and published promising results on the ImageNet dataset. Note that their method, named Xnor-Net, requires additional multiplication by a different scaling factor for each patch in each sample \citep{rastegari2016xnor} Section 3.2 Eq. 10 and figure 2). This in itself, requires many multiplications and prevents efficient implementation of XnorNet on known hardware designs. Moreover, \citep{rastegari2016xnor} didn't quantize first and last layers, therefore XNOR-Net are only partially binarized NNs. \cite{miyashita2016convolutional} suggested a more relaxed quantization (more than 1-bit) for both the weights and activation. Their idea was to quantize both and use shift operations as in our Eq.~(\ref{eq:straight-through-gradient}). They proposed to quantize the parameters in their non-uniform, base-2 logarithmic representation. This idea was inspired by the fact that the weights and activations in a trained network naturally have non-uniform distributions. They moreover showed that they can quantize the gradients as well to 6-bit without  significant losses in performance (on the Cifar-10 dataset). \citet{zhou2016dorefa} applied similar ideas to the ImageNet dataset and showed that by using 1-bit weights, 2-bit activations and 6-bit gradients they can achieve $46.1\%$ top-1 accuracies using the AlexNet architecture. They named this method DoReFa net. Here we outperform DoReFa net and achieve $46.8\%$ using a 1-2-6 bit quantization scheme (weight-activation-gradients) and $51\%$ using a 1-2-32 quantization scheme. These results confirm that we can achieve comparable results even on a large dataset by applying the Xnor kernel several times. \citet{merolla2016deep} showed that DNN can be robust to more than just weight binarization. They applied several different distortions to the weights, including additive and multiplicative noise, and a class of non-linear projections.This was shown to improve robustness to other distortions and even boost results. \citet{zhengbinarized} tried to apply our binarization scheme to recurrent neural network for language modeling and achieved comparable results as well. \citet{andri2016yodann} even created a hardware implementation to speed up BNNs. 

 

\section*{Conclusion}
 



We have introduced BNNs, which binarize deep neural networks and can lead to dramatic improvements in both power consumption and computation speed. During the forward pass (both at run-time and train-time), 
BNNs drastically reduce memory size and accesses, and replace most arithmetic operations with bit-wise operations. Our estimates indicate  that power efficiency can be improved by more than one order of magnitude (see Section \ref{sec:power}). In terms of speed, we programmed a binary matrix multiplication GPU kernel that enabled running MLP over the MNIST datset 7 times faster (than with an unoptimized GPU kernel) without any loss of accuracy (see Section \ref{sec:faster}).

We have shown that BNNs can handle MNIST, CIFAR-10 and SVHN while achieving nearly state-of-the-art accuracy. While our results for the challenging ImageNet are not on par with the best results achievable with full precision networks, they significantly improve all previous attempts to compress ImageNet-capable architectures. Moreover, by quantizing the weights and activations to more than 1-bit (i.e., QNNs), we have been able to achieve comparable results to the 32-bit floating point architectures (see Section~\ref{sec:preliminary} and supplementary material - Appendix B). A major open research avenue would be to further improve our results on ImageNet. Substantial progress in this direction might go a long way towards facilitating DNN usability in low power instruments such as mobile phones.  



\acks{We would like to express our appreciation to Elad Hoffer, for his technical assistance and constructive comments. We thank our fellow MILA lab members who took the time to read the article and give us some feedback. 
We thank the developers of Torch, \citep{Torch-2011} a Lua based environment, and Theano \citep{bergstra+al:2010-scipy,Bastien-Theano-2012}, a Python library that allowed us to easily develop fast and optimized code for GPU. 
We also thank the developers of Pylearn2 \citep{pylearn2_arxiv_2013} and Lasagne \citep{dieleman-et-al-2015}, two deep learning libraries built on the top of Theano. 
We thank Yuxin Wu for helping us compare our GPU kernels with cuBLAS.
 We are also grateful for funding from NSERC, the Canada Research Chairs, Compute Canada, and CIFAR.
We are also grateful for funding from CIFAR, NSERC, IBM, Samsung. This research was supported by The Israel Science Foundation (grant No. 1890/14)}


\newpage

\appendix

\section{Implementation Details}
In this section we give full implementation details over our  MNIST,SVHN, CIFAR-10 and ImageNet datasets. 
\subsection{MLP on MNIST (Theano)} 

MNIST is an image classification benchmark dataset \citep{LeCun+98}.
It consists of a training set of 60K and a test set of 10K 28 $\times$ 28 gray-scale images representing digits ranging from 0 to 9.
In order for this benchmark to remain a challenge, we did not use any convolution, data-augmentation, preprocessing or unsupervised learning.
The Multi-Layer-Perceptron (MLP) we train on MNIST consists of 3 hidden layers of 4096 binary units 
and a L2-SVM output layer; L2-SVM has been shown to perform 
better than Softmax on several classification benchmarks \citep{Tang-wkshp-2013,Lee-et-al-2014}.
We regularize the model with Dropout \citep{Srivastava14}.
The square hinge loss is minimized with the ADAM adaptive
learning rate method \citep{kingma2014adam}.
We use an exponentially decaying global learning rate, as per Algorithm 1,
and also scale the learning rates of the weights with their initialization coefficients from \citep{GlorotAISTATS2010-small}, 
as suggested by \citet{Courbariaux-et-al-2015}.
We use Batch Normalization with a minibatch of size 100 to speed up the training.
As is typical, we use the last 10K samples of the training set as a validation set for early stopping and model selection.
We report the test error rate associated with the best validation error rate after 1000 epochs
(we do not retrain on the validation set).

\subsection{MLP on MNIST (Torch7)} 

We use a similar architecture as in our Theano experiments, without dropout, and with 2048 binary units per layer instead of 4096. 
Additionally, we use the shift base AdaMax and BN (with a minibatch of size 100)  instead of the vanilla implementations, to reduce the number of multiplications. 
Likewise, we decay the learning rate by using a 1-bit right shift every 10 epochs.

\subsection{ConvNet on CIFAR-10 (Theano)}

CIFAR-10 is an image classification benchmark dataset.
It consists of a training set of size 50K and a test set of size 10K, where instances are 32 $\times$ 32 color images representing airplanes, automobiles, birds, cats, deer, dogs, frogs, horses, ships and trucks.
We do not use data-augmentation (which can really be a game changer for this dataset; see \citealt{Graham-2014}).
The architecture of our ConvNet is identical to that used by \citet{Courbariaux2015} except for the binarization of the activations.
The \citet{Courbariaux-et-al-2015} architecture is itself mainly inspired by VGG \citep{Simonyan2015}.
The square hinge loss is minimized with ADAM.
We use an exponentially decaying learning rate, as we did for MNIST.
We scale the learning rates of the weights with their initialization coefficients from \citep{GlorotAISTATS2010-small}.
We use Batch Normalization with a minibatch of size 50 to speed up the training.
We use the last 5000 samples of the training set as a validation set.
We report the test error rate associated with the best validation error rate after 500 training epochs
(we do not retrain on the validation set).

\begin{table}[h]
\caption{Architecture of our CIFAR-10 ConvNet. We only use "same" convolutions as in VGG \citep{Simonyan2015}.}
\begin{center}
\scalebox{.75}{
\begin{tabular}{@{}l@{}}
\toprule
CIFAR-10 ConvNet architecture                  \\ \midrule
Input: $32\times32$ - RGB image                         \\
$3\times3$ - 128 convolution layer                      \\
BatchNorm and Binarization layers              \\
$3\times3$ - 128 convolution and $2\times2$ max-pooling layers \\
BatchNorm and Binarization layers              \\ \midrule
$3\times3$ - 256 convolution layer                      \\
BatchNorm and Binarization layers              \\
$3\times3$ - 256 convolution and $2\times2$ max-pooling layers \\
BatchNorm and Binarization layers              \\ \midrule
$3\times3$ - 512 convolution layer                      \\
BatchNorm and Binarization layers              \\
$3\times3$ - 512 convolution and $2\times2$ max-pooling layers \\
BatchNorm and Binarization layers              \\ \midrule
1024 fully connected layer                     \\
BatchNorm and Binarization layers              \\
1024 fully connected layer                     \\
BatchNorm and Binarization layers              \\ \midrule
10 fully connected layer                       \\ 
BatchNorm layer (no binarization)              \\
Cost: Mean square hinge loss                   \\ \bottomrule
\end{tabular}
}
\end{center}
\label{tab:architecture}
\end{table}

\subsection{ConvNet on CIFAR-10 (Torch7)} 

We use the same architecture as in our Theano experiments. 
We apply shift-based AdaMax and BN (with a minibatch of size 200) instead of the vanilla implementations to reduce the number of multiplications. 
Likewise, we decay the learning rate by using a 1-bit right shift every 50 epochs.

\subsection{ConvNet on SVHN}

SVHN is also an image classification benchmark dataset.
It consists of a training set of size 604K examples and a test set of size 26K, where instances are 32 $\times$ 32 color images representing digits ranging from 0 to 9.
In both sets of experiments, we follow the same procedure used for the CIFAR-10 experiments, 
with a few notable exceptions: we use half the number of units in the convolution layers, and we train for 200 epochs instead of 500 
(because SVHN is a much larger dataset than CIFAR-10).

\subsection{ConvNet on ImageNet}
\label{appendix-ImageNet}
ImageNet classification task  consists of a training set of size 1.2M samples and test set of size 50K. Each instance is labeled with one of 1000 categories including objects, animals, scenes, and even some abstract shapes.
\paragraph{AlexNet:} Our AlexNet implementation consists of 5 convolution layers followed by 3 fully connected layers (see Section  \ref{tab:AlexNet architecture}). Additionally, we use  Adam as our optimization method and batch-normalization layers (with a minibatch of size 512). Likewise, we decay the learning rate by 0.1 every 20 epochs. 
\paragraph{GoogleNet:} Our GoogleNet implementation consist of 2 convolution layers followed by 10 inception layers, spatial-average-pooling and a fully connected classifier. We also used the 2 auxilary classifiers. Additionally, we use  Adam \citep{Kingma2015} as our optimization method and batch-normalization layers (with a minibatch of size 64). Likewise, we decay the learning rate by 0.1 every 10 epochs.


\begin{table}[ht]
\caption{Our AlexNet Architecture.}
\begin{center}
\scalebox{.75}{
\begin{tabular}{@{}l@{}}
\toprule
AlexNet ConvNet architecture                  \\ \midrule
Input: $32\times32$ - RGB image                         \\
$11\times11$ - 64 convolution layer and $3\times3$ max-pooling layers   \\
BatchNorm and Binarization layers              \\
$5\times5$ - 192 convolution layer   and $3\times3$ max-pooling layers  \\
BatchNorm and Binarization layers              \\
$3\times3$ - 384 convolution layer                      \\
BatchNorm and Binarization layers              \\
$3\times3$ - 256 convolution layer  \\
BatchNorm and Binarization layers              \\ 
$3\times3$ - 256 convolution layer                      \\
BatchNorm and Binarization layers              \\
4096 fully connected layer                     \\
BatchNorm and Binarization layers              \\
4096 fully connected layer                     \\
BatchNorm and Binarization layers              \\ \midrule
1000 fully connected layer                       \\ 
BatchNorm layer (no binarization) \\
SoftMax layer (no binarization) \\
Cost: Negative log likelihood                   \\ \bottomrule
\end{tabular}
}
\end{center}
\label{tab:AlexNet architecture}
\end{table}

\vskip 0.2in
\bibliography{sample}

\begin{thebibliography}{74}
\providecommand{\natexlab}[1]{#1}
\providecommand{\url}[1]{\texttt{#1}}
\expandafter\ifx\csname urlstyle\endcsname\relax
  \providecommand{\doi}[1]{doi: #1}\else
  \providecommand{\doi}{doi: \begingroup \urlstyle{rm}\Url}\fi

\bibitem[Andri et~al.(2016)Andri, Cavigelli, Rossi, and
  Benini]{andri2016yodann}
Renzo Andri, Lukas Cavigelli, Davide Rossi, and Luca Benini.
\newblock Yodann: An ultra-low power convolutional neural network accelerator
  based on binary weights.
\newblock \emph{arXiv preprint arXiv:1606.05487}, 2016.

\bibitem[Bahdanau et~al.(2015)Bahdanau, Cho, and
  Bengio]{Bahdanau-et-al-ICLR2015-small}
Dzmitry Bahdanau, Kyunghyun Cho, and Yoshua Bengio.
\newblock Neural machine translation by jointly learning to align and
  translate.
\newblock In \emph{ICLR'2015, arXiv:1409.0473}, 2015.

\bibitem[Baldassi et~al.(2015)Baldassi, Ingrosso, Lucibello, Saglietti, and
  Zecchina]{Baldassi2015}
Carlo Baldassi, Alessandro Ingrosso, Carlo Lucibello, Luca Saglietti, and
  Riccardo Zecchina.
\newblock {Subdominant Dense Clusters Allow for Simple Learning and High
  Computational Performance in Neural Networks with Discrete Synapses}.
\newblock \emph{Physical Review Letters}, 115\penalty0 (12):\penalty0 1--5,
  2015.
\newblock ISSN 10797114.
\newblock \doi{10.1103/PhysRevLett.115.128101}.

\bibitem[Bastien et~al.(2012)Bastien, Lamblin, Pascanu, Bergstra, Goodfellow,
  Bergeron, Bouchard, and Bengio]{Bastien-Theano-2012}
Fr{\'{e}}d{\'{e}}ric Bastien, Pascal Lamblin, Razvan Pascanu, James Bergstra,
  Ian~J. Goodfellow, Arnaud Bergeron, Nicolas Bouchard, and Yoshua Bengio.
\newblock Theano: new features and speed improvements.
\newblock Deep Learning and Unsupervised Feature Learning NIPS 2012 Workshop,
  2012.

\bibitem[Beauchamp et~al.(2006)Beauchamp, Hauck, Underwood, and
  Hemmert]{Beauchamp-et-al-2006}
Michael~J Beauchamp, Scott Hauck, Keith~D Underwood, and K~Scott Hemmert.
\newblock Embedded floating-point units in {FPGAs}.
\newblock In \emph{Proceedings of the 2006 ACM/SIGDA 14th international
  symposium on Field programmable gate arrays}, pages 12--20. ACM, 2006.

\bibitem[Bengio(2013)]{Bengio-arxiv2013}
Yoshua Bengio.
\newblock Estimating or propagating gradients through stochastic neurons.
\newblock Technical Report arXiv:1305.2982, Universite de Montreal, 2013.

\bibitem[Bergstra et~al.(2010)Bergstra, Breuleux, Bastien, Lamblin, Pascanu,
  Desjardins, Turian, Warde-Farley, and Bengio]{bergstra+al:2010-scipy}
James Bergstra, Olivier Breuleux, Fr{\'{e}}d{\'{e}}ric Bastien, Pascal Lamblin,
  Razvan Pascanu, Guillaume Desjardins, Joseph Turian, David Warde-Farley, and
  Yoshua Bengio.
\newblock Theano: a {CPU} and {GPU} math expression compiler.
\newblock In \emph{Proceedings of the Python for Scientific Computing
  Conference ({SciPy})}, June 2010.
\newblock Oral Presentation.

\bibitem[Chen et~al.(2014{\natexlab{a}})Chen, Du, Sun, Wang, Wu, Chen, and
  Temam]{Chen-et-al-ACM2014}
Tianshi Chen, Zidong Du, Ninghui Sun, Jia Wang, Chengyong Wu, Yunji Chen, and
  Olivier Temam.
\newblock Diannao: A small-footprint high-throughput accelerator for ubiquitous
  machine-learning.
\newblock In \emph{Proceedings of the 19th international conference on
  Architectural support for programming languages and operating systems}, pages
  269--284. ACM, 2014{\natexlab{a}}.

\bibitem[Chen et~al.(2015)Chen, Wilson, Tyree, Weinberger, and
  Chen]{chen2015compressing}
Wenlin Chen, James~T Wilson, Stephen Tyree, Kilian~Q Weinberger, and Yixin
  Chen.
\newblock Compressing neural networks with the hashing trick.
\newblock \emph{arXiv preprint arXiv:1504.04788}, 2015.

\bibitem[Chen et~al.(2014{\natexlab{b}})Chen, Luo, Liu, Zhang, He, Wang, Li,
  Chen, Xu, Sun, et~al.]{Chen-et-al-IEEE2014}
Yunji Chen, Tao Luo, Shaoli Liu, Shijin Zhang, Liqiang He, Jia Wang, Ling Li,
  Tianshi Chen, Zhiwei Xu, Ninghui Sun, et~al.
\newblock Dadiannao: A machine-learning supercomputer.
\newblock In \emph{Microarchitecture (MICRO), 2014 47th Annual IEEE/ACM
  International Symposium on}, pages 609--622. IEEE, 2014{\natexlab{b}}.

\bibitem[Cheng et~al.(2015)Cheng, Soudry, Mao, and Lan]{Cheng-et-al-2015}
Zhiyong Cheng, Daniel Soudry, Zexi Mao, and Zhenzhong Lan.
\newblock Training binary multilayer neural networks for image classification
  using expectation backpropgation.
\newblock \emph{arXiv preprint arXiv:1503.03562}, 2015.

\bibitem[Coates et~al.(2013)Coates, Huval, Wang, Wu, Catanzaro, and
  Andrew]{Coates-et-al-2013}
Adam Coates, Brody Huval, Tao Wang, David Wu, Bryan Catanzaro, and Ng~Andrew.
\newblock Deep learning with {COTS} {HPC} systems.
\newblock In \emph{Proceedings of the 30th international conference on machine
  learning}, pages 1337--1345, 2013.

\bibitem[Collobert et~al.(2011)Collobert, Kavukcuoglu, and Farabet]{Torch-2011}
Ronan Collobert, Koray Kavukcuoglu, and Cl\'{e}ment Farabet.
\newblock Torch7: A matlab-like environment for machine learning.
\newblock In \emph{Big{L}earn, {NIPS} {W}orkshop}, 2011.

\bibitem[Courbariaux et~al.(2014)Courbariaux, Bengio, and
  David]{courbariaux+al-TR2014}
Matthieu Courbariaux, Yoshua Bengio, and Jean-Pierre David.
\newblock Training deep neural networks with low precision multiplications.
\newblock \emph{ArXiv e-prints}, abs/1412.7024, December 2014.

\bibitem[Courbariaux et~al.(2015{\natexlab{a}})Courbariaux, Bengio, and
  David]{Courbariaux-et-al-2015}
Matthieu Courbariaux, Yoshua Bengio, and Jean-Pierre David.
\newblock Binaryconnect: Training deep neural networks with binary weights
  during propagations.
\newblock \emph{ArXiv e-prints}, abs/1511.00363, November 2015{\natexlab{a}}.

\bibitem[Courbariaux et~al.(2015{\natexlab{b}})Courbariaux, Bengio, and
  David]{Courbariaux2015}
Matthieu Courbariaux, Yoshua Bengio, and Jean-Pierre David.
\newblock {BinaryConnect: Training Deep Neural Networks with binary weights
  during propagations}.
\newblock \emph{Nips}, pages 1--9, 2015{\natexlab{b}}.
\newblock URL \url{http://arxiv.org/abs/1511.00363}.

\bibitem[Devlin et~al.(2014)Devlin, Zbib, Huang, Lamar, Schwartz, and
  Makhoul]{Devlin-et-al-ACL2014}
Jacob Devlin, Rabih Zbib, Zhongqiang Huang, Thomas Lamar, Richard Schwartz, and
  John Makhoul.
\newblock Fast and robust neural network joint models for statistical machine
  translation.
\newblock In \emph{Proc. ACL'2014}, 2014.

\bibitem[Dieleman et~al.(2015)Dieleman, Schlüter, Raffel, Olson, Sønderby,
  Nouri, Maturana, Thoma, Battenberg, Kelly, Fauw, Heilman, diogo149, McFee,
  Weideman, takacsg84, peterderivaz, Jon, instagibbs, Rasul, CongLiu,
  Britefury, and Degrave]{dieleman-et-al-2015}
Sander Dieleman, Jan Schlüter, Colin Raffel, Eben Olson, Søren~Kaae
  Sønderby, Daniel Nouri, Daniel Maturana, Martin Thoma, Eric Battenberg, Jack
  Kelly, Jeffrey~De Fauw, Michael Heilman, diogo149, Brian McFee, Hendrik
  Weideman, takacsg84, peterderivaz, Jon, instagibbs, Dr.~Kashif Rasul,
  CongLiu, Britefury, and Jonas Degrave.
\newblock Lasagne: First release., August 2015.
\newblock URL \url{http://dx.doi.org/10.5281/zenodo.27878}.

\bibitem[Esser et~al.(2015)Esser, Appuswamy, Merolla, Arthur, and
  Modha]{Esser-et-al-2015}
Steve~K Esser, Rathinakumar Appuswamy, Paul Merolla, John~V Arthur, and
  Dharmendra~S Modha.
\newblock Backpropagation for energy-efficient neuromorphic computing.
\newblock In \emph{Advances in Neural Information Processing Systems}, pages
  1117--1125, 2015.

\bibitem[Farabet et~al.(2011{\natexlab{a}})Farabet, LeCun, Kavukcuoglu,
  Culurciello, Martini, Akselrod, and Talay]{Farabet-et-al-2011-a}
Cl{\'e}ment Farabet, Yann LeCun, Koray Kavukcuoglu, Eugenio Culurciello, Berin
  Martini, Polina Akselrod, and Selcuk Talay.
\newblock Large-scale {FPGA-based} convolutional networks.
\newblock \emph{Machine Learning on Very Large Data Sets}, 1,
  2011{\natexlab{a}}.

\bibitem[Farabet et~al.(2011{\natexlab{b}})Farabet, Martini, Corda, Akselrod,
  Culurciello, and LeCun]{Farabet-et-al-2011-b}
Cl{\'e}ment Farabet, Berin Martini, Benoit Corda, Polina Akselrod, Eugenio
  Culurciello, and Yann LeCun.
\newblock Neuflow: A runtime reconfigurable dataflow processor for vision.
\newblock In \emph{Computer Vision and Pattern Recognition Workshops (CVPRW),
  2011 IEEE Computer Society Conference on}, pages 109--116. IEEE,
  2011{\natexlab{b}}.

\bibitem[Glorot and Bengio(2010)]{GlorotAISTATS2010-small}
Xavier Glorot and Yoshua Bengio.
\newblock Understanding the difficulty of training deep feedforward neural
  networks.
\newblock In \emph{AISTATS'2010}, 2010.

\bibitem[Gong et~al.(2014)Gong, Liu, Yang, and Bourdev]{Gong-et-al-2014}
Yunchao Gong, Liu Liu, Ming Yang, and Lubomir Bourdev.
\newblock Compressing deep convolutional networks using vector quantization.
\newblock \emph{arXiv preprint arXiv:1412.6115}, 2014.

\bibitem[Goodfellow et~al.(2013{\natexlab{a}})Goodfellow, Warde-Farley,
  Lamblin, Dumoulin, Mirza, Pascanu, Bergstra, Bastien, and
  Bengio]{pylearn2_arxiv_2013}
Ian~J. Goodfellow, David Warde-Farley, Pascal Lamblin, Vincent Dumoulin, Mehdi
  Mirza, Razvan Pascanu, James Bergstra, Fr{\'{e}}d{\'{e}}ric Bastien, and
  Yoshua Bengio.
\newblock Pylearn2: a machine learning research library.
\newblock \emph{arXiv preprint arXiv:1308.4214}, 2013{\natexlab{a}}.

\bibitem[Goodfellow et~al.(2013{\natexlab{b}})Goodfellow, Warde-Farley, Mirza,
  Courville, and Bengio]{Goodfellow2013a}
Ian~J. Goodfellow, David Warde-Farley, Mehdi Mirza, Aaron Courville, and Yoshua
  Bengio.
\newblock {Maxout Networks}.
\newblock \emph{arXiv preprint}, pages 1319--1327, 2013{\natexlab{b}}.
\newblock URL \url{http://arxiv.org/abs/1302.4389}.

\bibitem[Govindu et~al.(2004)Govindu, Zhuo, Choi, and
  Prasanna]{Govindu-et-al-2004}
Gokul Govindu, Ling Zhuo, Seonil Choi, and Viktor Prasanna.
\newblock Analysis of high-performance floating-point arithmetic on {FPGAs}.
\newblock In \emph{Parallel and Distributed Processing Symposium, 2004.
  Proceedings. 18th International}, page 149. IEEE, 2004.

\bibitem[Graham(2014)]{Graham-2014}
Benjamin Graham.
\newblock Spatially-sparse convolutional neural networks.
\newblock \emph{arXiv preprint arXiv:1409.6070}, 2014.

\bibitem[Graves(2011)]{Graves-2011-practical}
Alex Graves.
\newblock Practical variational inference for neural networks.
\newblock In \emph{Advances in Neural Information Processing Systems}, pages
  2348--2356, 2011.

\bibitem[Gupta et~al.(2015)Gupta, Agrawal, Gopalakrishnan, and
  Narayanan]{gupta2015deep}
Suyog Gupta, Ankur Agrawal, Kailash Gopalakrishnan, and Pritish Narayanan.
\newblock Deep learning with limited numerical precision.
\newblock \emph{CoRR, abs/1502.02551}, 392, 2015.

\bibitem[Gysel et~al.(2016)Gysel, Motamedi, and Ghiasi]{gysel2016hardware}
Philipp Gysel, Mohammad Motamedi, and Soheil Ghiasi.
\newblock Hardware-oriented approximation of convolutional neural networks.
\newblock \emph{arXiv preprint arXiv:1604.03168}, 2016.

\bibitem[Han and Dally(2015)]{Han2015}
Huizi~Mao Han, Song and William~J. Dally.
\newblock {Deep Compression: Compressing Deep Neural Networks with Pruning,
  Trained Quantization and Huffman Coding}.
\newblock \emph{arXiv preprint}, pages 1--11, 2015.
\newblock URL \url{http://arxiv.org/abs/1510.00149}.

\bibitem[Han et~al.(2015{\natexlab{a}})Han, Mao, and Dally]{han2015deep}
Song Han, Huizi Mao, and William~J Dally.
\newblock Deep compression: Compressing deep neural networks with pruning,
  trained quantization and huffman coding.
\newblock \emph{arXiv preprint arXiv:1510.00149}, 2015{\natexlab{a}}.

\bibitem[Han et~al.(2015{\natexlab{b}})Han, Pool, Tran, and
  Dally]{Han-et-al-2015}
Song Han, Jeff Pool, John Tran, and William Dally.
\newblock Learning both weights and connections for efficient neural network.
\newblock In \emph{Advances in Neural Information Processing Systems}, pages
  1135--1143, 2015{\natexlab{b}}.

\bibitem[Hinton(2012)]{Hinton-Coursera2012}
Geoffrey Hinton.
\newblock Neural networks for machine learning.
\newblock Coursera, video lectures, 2012.

\bibitem[Hinton et~al.(2012)Hinton, Deng, Dahl, Mohamed, Jaitly, Senior,
  Vanhoucke, Nguyen, Sainath, and Kingsbury]{Hinton-et-al-2012}
Geoffrey Hinton, Li~Deng, George~E. Dahl, {Abdel-rahman} Mohamed, Navdeep
  Jaitly, Andrew Senior, Vincent Vanhoucke, Patrick Nguyen, Tara Sainath, and
  Brian Kingsbury.
\newblock Deep neural networks for acoustic modeling in speech recognition.
\newblock \emph{{IEEE} Signal Processing Magazine}, 29\penalty0 (6):\penalty0
  82--97, Nov. 2012.

\bibitem[Hochreiter and Schmidhuber(1997)]{hochreiter1997long}
Sepp Hochreiter and J{\"u}rgen Schmidhuber.
\newblock Long short-term memory.
\newblock \emph{Neural computation}, 9\penalty0 (8):\penalty0 1735--1780, 1997.

\bibitem[Horowitz(2014)]{Horowitz2014}
Mark Horowitz.
\newblock {Computing's Energy Problem (and what we can do about it)}.
\newblock \emph{IEEE Interational Solid State Circuits Conference}, pages
  10--14, 2014.
\newblock ISSN 0018-9200.
\newblock \doi{10.1109/JSSC.2014.2361354}.

\bibitem[Hwang and Sung(2014)]{hwang-et-al-2014}
Kyuyeon Hwang and Wonyong Sung.
\newblock Fixed-point feedforward deep neural network design using weights+ 1,
  0, and- 1.
\newblock In \emph{Signal Processing Systems (SiPS), 2014 IEEE Workshop on},
  pages 1--6. IEEE, 2014.

\bibitem[Ioffe and Szegedy(2015)]{Ioffe+Szegedy-2015}
Sergey Ioffe and Christian Szegedy.
\newblock Batch normalization: Accelerating deep network training by reducing
  internal covariate shift.
\newblock 2015.

\bibitem[{Kim} and {Smaragdis}(2016)]{Kim-et-al-2016}
M.~{Kim} and P.~{Smaragdis}.
\newblock {Bitwise Neural Networks}.
\newblock \emph{ArXiv e-prints}, January 2016.

\bibitem[Kingma and Ba(2014{\natexlab{a}})]{Kingma2015}
Diederik Kingma and Jimmy Ba.
\newblock {Adam: A Method for Stochastic Optimization}.
\newblock \emph{arXiv:1412.6980 [cs]}, pages 1--13, 2014{\natexlab{a}}.
\newblock URL
  \url{http://arxiv.org/abs/1412.6980$\backslash$nhttp://www.arxiv.org/pdf/1412.6980.pdf}.

\bibitem[Kingma and Ba(2014{\natexlab{b}})]{kingma2014adam}
Diederik Kingma and Jimmy Ba.
\newblock Adam: A method for stochastic optimization.
\newblock \emph{arXiv preprint arXiv:1412.6980}, 2014{\natexlab{b}}.

\bibitem[Krizhevsky et~al.(2012)Krizhevsky, Sutskever, and
  Hinton]{Krizhevsky-2012-small}
A.~Krizhevsky, I.~Sutskever, and G.~Hinton.
\newblock {ImageNet} classification with deep convolutional neural networks.
\newblock In \emph{NIPS'2012}. 2012.

\bibitem[{LeCun} et~al.(1998){LeCun}, Bottou, Bengio, and Haffner]{LeCun+98}
Yann {LeCun}, Leon Bottou, Yoshua Bengio, and Patrick Haffner.
\newblock Gradient-based learning applied to document recognition.
\newblock \emph{Proceedings of the {IEEE}}, 86\penalty0 (11):\penalty0
  2278--2324, November 1998.

\bibitem[Lee et~al.(2014)Lee, Xie, Gallagher, Zhang, and Tu]{Lee-et-al-2014}
Chen-Yu Lee, Saining Xie, Patrick Gallagher, Zhengyou Zhang, and Zhuowen Tu.
\newblock Deeply-supervised nets.
\newblock \emph{arXiv preprint arXiv:1409.5185}, 2014.

\bibitem[Lee et~al.(2015)Lee, Gallagher, and Tu]{lee-et-al-2015}
Chen-Yu Lee, Patrick~W Gallagher, and Zhuowen Tu.
\newblock Generalizing pooling functions in convolutional neural networks:
  Mixed, gated, and tree.
\newblock \emph{arXiv preprint arXiv:1509.08985}, 2015.

\bibitem[Lin et~al.(2015{\natexlab{a}})Lin, Courbariaux, Memisevic, and
  Bengio]{Lin-et-al-2015}
Zhouhan Lin, Matthieu Courbariaux, Roland Memisevic, and Yoshua Bengio.
\newblock Neural networks with few multiplications.
\newblock \emph{ArXiv e-prints}, abs/1510.03009, October 2015{\natexlab{a}}.

\bibitem[Lin et~al.(2015{\natexlab{b}})Lin, Courbariaux, Memisevic, and
  Bengio]{Lin2015}
Zhouhan Lin, Matthieu Courbariaux, Roland Memisevic, and Yoshua Bengio.
\newblock {Neural Networks with Few Multiplications}.
\newblock \emph{Iclr}, pages 1--8, 2015{\natexlab{b}}.
\newblock URL \url{http://arxiv.org/abs/1510.03009}.

\bibitem[Lomont(2003)]{lomont2003fast}
Chris Lomont.
\newblock Fast inverse square root.
\newblock \emph{Tech-315 nical Report}, page~32, 2003.

\bibitem[Marcus et~al.(1993)Marcus, Marcinkiewicz, and
  Santorini]{marcus1993building}
Mitchell~P Marcus, Mary~Ann Marcinkiewicz, and Beatrice Santorini.
\newblock Building a large annotated corpus of english: The penn treebank.
\newblock \emph{Computational linguistics}, 19\penalty0 (2):\penalty0 313--330,
  1993.

\bibitem[Merolla et~al.(2016)Merolla, Appuswamy, Arthur, Esser, and
  Modha]{merolla2016deep}
Paul Merolla, Rathinakumar Appuswamy, John Arthur, Steve~K Esser, and
  Dharmendra Modha.
\newblock Deep neural networks are robust to weight binarization and other
  non-linear distortions.
\newblock \emph{arXiv preprint arXiv:1606.01981}, 2016.

\bibitem[Mikolov and Zweig(2012)]{mikolov2012context}
Tomas Mikolov and Geoffrey Zweig.
\newblock Context dependent recurrent neural network language model.
\newblock In \emph{SLT}, pages 234--239, 2012.

\bibitem[Miyashita et~al.(2016)Miyashita, Lee, and
  Murmann]{miyashita2016convolutional}
Daisuke Miyashita, Edward~H Lee, and Boris Murmann.
\newblock Convolutional neural networks using logarithmic data representation.
\newblock \emph{arXiv preprint arXiv:1603.01025}, 2016.

\bibitem[Mnih et~al.(2015)Mnih, Kavukcuoglo, Silver, Rusu, Veness, Bellemare,
  Graves, Riedmiller, Fidgeland, Ostrovski, Petersen, Beattie, Sadik,
  Antonoglou, King, Kumaran, Wierstra, Legg, and Hassabis]{Mnih-et-al-2015}
Volodymyr Mnih, Koray Kavukcuoglo, David Silver, Andrei~A. Rusu, Joel Veness,
  Marc~G. Bellemare, Alex Graves, Martin Riedmiller, Andreas~K. Fidgeland,
  Georg Ostrovski, Stig Petersen, Charles Beattie, Amir Sadik, Ioannis
  Antonoglou, Helen King, Dharsan Kumaran, Daan Wierstra, Shane Legg, and Demis
  Hassabis.
\newblock Human-level control through deep reinforcement learning.
\newblock \emph{Nature}, 518:\penalty0 529--533, 2015.

\bibitem[Mordvintsev et~al.(2015)Mordvintsev, Olah, and
  Tyka]{Mordvintsev-et-al-2015}
Alexander Mordvintsev, Christopher Olah, and Mike Tyka.
\newblock Inceptionism: Going deeper into neural networks, 2015.
\newblock URL
  \url{http://googleresearch.blogspot.co.uk/2015/06/inceptionism-going-deeper-into-neural.html}.
\newblock Accessed: 2015-06-30.

\bibitem[Ott et~al.(2016)Ott, Lin, Zhang, Liu, and Bengio]{ott2016recurrent}
Joachim Ott, Zhouhan Lin, Ying Zhang, Shih-Chii Liu, and Yoshua Bengio.
\newblock Recurrent neural networks with limited numerical precision.
\newblock \emph{arXiv preprint arXiv:1608.06902}, 2016.

\bibitem[Pham et~al.(2012)Pham, Jelaca, Farabet, Martini, LeCun, and
  Culurciello]{Pham-et-al-2012}
Phi-Hung Pham, Darko Jelaca, Clement Farabet, Berin Martini, Yann LeCun, and
  Eugenio Culurciello.
\newblock Neuflow: Dataflow vision processing system-on-a-chip.
\newblock In \emph{Circuits and Systems (MWSCAS), 2012 IEEE 55th International
  Midwest Symposium on}, pages 1044--1047. IEEE, 2012.

\bibitem[Rastegari et~al.(2016)Rastegari, Ordonez, Redmon, and
  Farhadi]{rastegari2016xnor}
Mohammad Rastegari, Vicente Ordonez, Joseph Redmon, and Ali Farhadi.
\newblock Xnor-net: Imagenet classification using binary convolutional neural
  networks.
\newblock \emph{arXiv preprint arXiv:1603.05279}, 2016.

\bibitem[Romero et~al.(2014)Romero, Ballas, Kahou, Chassang, Gatta, and
  Bengio]{Romero-et-al-2014}
Adriana Romero, Nicolas Ballas, Samira~Ebrahimi Kahou, Antoine Chassang, Carlo
  Gatta, and Yoshua Bengio.
\newblock Fitnets: Hints for thin deep nets.
\newblock \emph{arXiv preprint arXiv:1412.6550}, 2014.

\bibitem[Sainath et~al.(2013)Sainath, rahman Mohamed, Kingsbury, and
  Ramabhadran]{Sainath-et-al-ICASSP2013}
Tara Sainath, Abdel rahman Mohamed, Brian Kingsbury, and Bhuvana Ramabhadran.
\newblock Deep convolutional neural networks for {LVCSR}.
\newblock In \emph{ICASSP 2013}, 2013.

\bibitem[Silver et~al.(2016)Silver, Huang, Maddison, Guez, Sifre, van~den
  Driessche, Schrittwieser, Antonoglou, Panneershelvam, Lanctot, Dieleman,
  Grewe, Nham, Kalchbrenner, Sutskever, Lillicrap, Leach, Kavukcuoglu, Graepel,
  and Hassabis]{Silver-et-al-2016}
David Silver, Aja Huang, Chris~J. Maddison, Arthur Guez, Laurent Sifre, George
  van~den Driessche, Julian Schrittwieser, Ioannis Antonoglou, Veda
  Panneershelvam, Marc Lanctot, Sander Dieleman, Dominik Grewe, John Nham, Nal
  Kalchbrenner, Ilya Sutskever, Timothy Lillicrap, Madeleine Leach, Koray
  Kavukcuoglu, Thore Graepel, and Demis Hassabis.
\newblock Mastering the game of go with deep neural networks and tree search.
\newblock \emph{Nature}, 529\penalty0 (7587):\penalty0 484--489, Jan 2016.
\newblock ISSN 0028-0836.
\newblock URL \url{http://dx.doi.org/10.1038/nature16961}.
\newblock Article.

\bibitem[Simonyan and Zisserman(2015)]{Simonyan2015}
Karen Simonyan and Andrew Zisserman.
\newblock Very deep convolutional networks for large-scale image recognition.
\newblock In \emph{ICLR}, 2015.

\bibitem[Soudry et~al.(2014)Soudry, Hubara, and
  Meir]{Soudry-et-al-NIPS2014-small}
Daniel Soudry, Itay Hubara, and Ron Meir.
\newblock Expectation backpropagation: Parameter-free training of multilayer
  neural networks with continuous or discrete weights.
\newblock In \emph{NIPS'2014}, 2014.

\bibitem[Spang and Schultheiss(1962)]{spang1962reduction}
H~Spang and P~Schultheiss.
\newblock Reduction of quantizing noise by use of feedback.
\newblock \emph{IRE Transactions on Communications Systems}, 10\penalty0
  (4):\penalty0 373--380, 1962.

\bibitem[Srivastava et~al.(2014)Srivastava, Hinton, Krizhevsky, Sutskever, and
  Salakhutdinov]{Srivastava14}
Nitish Srivastava, Geoffrey Hinton, Alex Krizhevsky, Ilya Sutskever, and Ruslan
  Salakhutdinov.
\newblock Dropout: A simple way to prevent neural networks from overfitting.
\newblock \emph{Journal of Machine Learning Research}, 15:\penalty0 1929--1958,
  2014.
\newblock URL \url{http://jmlr.org/papers/v15/srivastava14a.html}.

\bibitem[Sutskever et~al.(2014)Sutskever, Vinyals, and
  Le]{Sutskever-et-al-NIPS2014}
Ilya Sutskever, Oriol Vinyals, and Quoc~V. Le.
\newblock Sequence to sequence learning with neural networks.
\newblock In \emph{NIPS'2014}, 2014.

\bibitem[Szegedy et~al.(2014)Szegedy, Liu, Jia, Sermanet, Reed, Anguelov,
  Erhan, Vanhoucke, and Rabinovich]{Szegedy-et-al-arxiv2014}
Christian Szegedy, Wei Liu, Yangqing Jia, Pierre Sermanet, Scott Reed, Dragomir
  Anguelov, Dumitru Erhan, Vincent Vanhoucke, and Andrew Rabinovich.
\newblock Going deeper with convolutions.
\newblock Technical report, arXiv:1409.4842, 2014.

\bibitem[Tang(2013)]{Tang-wkshp-2013}
Yichuan Tang.
\newblock Deep learning using linear support vector machines.
\newblock Workshop on Challenges in Representation Learning, {ICML}, 2013.

\bibitem[Torii et~al.(2016)Torii, Kokubo, Yamamoto, Itoh, Takenaka, and
  Matsumoto]{torii2016asic}
Naoya Torii, Hirotaka Kokubo, Dai Yamamoto, Kouichi Itoh, Masahiko Takenaka,
  and Tsutomu Matsumoto.
\newblock Asic implementation of random number generators using sr latches and
  its evaluation.
\newblock \emph{EURASIP Journal on Information Security}, 2016\penalty0
  (1):\penalty0 1--12, 2016.

\bibitem[Vanhoucke et~al.(2011)Vanhoucke, Senior, and
  Mao]{Vanhoucke-et-al-2011}
Vincent Vanhoucke, Andrew Senior, and Mark~Z Mao.
\newblock Improving the speed of neural networks on {CPUs}.
\newblock In \emph{Proc. Deep Learning and Unsupervised Feature Learning NIPS
  Workshop}, 2011.

\bibitem[Wan et~al.(2013)Wan, Zeiler, Zhang, LeCun, and
  Fergus]{Wan+al-ICML2013-small}
Li~Wan, Matthew Zeiler, Sixin Zhang, Yann LeCun, and Rob Fergus.
\newblock Regularization of neural networks using dropconnect.
\newblock In \emph{ICML'2013}, 2013.

\bibitem[Zhang et~al.(2015)Zhang, Zou, Ming, He, and Sun]{zhang2015efficient}
Xiangyu Zhang, Jianhua Zou, Xiang Ming, Kaiming He, and Jian Sun.
\newblock Efficient and accurate approximations of nonlinear convolutional
  networks.
\newblock pages 1984--1992, 2015.

\bibitem[Zheng and Tang()]{zhengbinarized}
Weiyi Zheng and Yina Tang.
\newblock Binarized neural networks for language modeling.

\bibitem[Zhou et~al.(2016)Zhou, Ni, Zhou, Wen, Wu, and Zou]{zhou2016dorefa}
Shuchang Zhou, Zekun Ni, Xinyu Zhou, He~Wen, Yuxin Wu, and Yuheng Zou.
\newblock Dorefa-net: Training low bitwidth convolutional neural networks with
  low bitwidth gradients.
\newblock \emph{arXiv preprint arXiv:1606.06160}, 2016.

\end{thebibliography}

\end{document}